%% file: neurips_2026.tex
\documentclass{article}

\usepackage[preprint]{neurips_2026}

\usepackage[utf8]{inputenc} 
\usepackage[T1]{fontenc}    
\usepackage{hyperref}       
\usepackage{url}            
\usepackage{booktabs}       
\usepackage{amsfonts}       
\usepackage{nicefrac}       
\usepackage{microtype}      
\usepackage{xcolor}         
\usepackage{algorithm}     
\usepackage{algpseudocode}
\usepackage{amsmath}
\usepackage{amssymb}
\usepackage{mathtools}
\usepackage{amsthm}
\usepackage{tikz}
\usepackage{float}
\usepackage{listings}
\usepackage{xcolor}
\usepackage[framemethod=TikZ]{mdframed}
\usetikzlibrary{shapes.geometric, arrows}

\title{eMoT: evolving Memory-of-Thought via Symbolic Anchoring and Memory Corrosion}

\author{%
  {\bf Xiang Li}, {\bf Jiwei Wei}\thanks{Corresponding author.}, {\bf Ke Liu}, {\bf Yitong Qin}, \\
  {\bf Jinyu Guo}, {\bf Malu Zhang}, {\bf Peng Wang}, {\bf Yang Yang} \\
  Center for Future Media, \\
  University of Electronic Science and Technology of China, \\
  Chengdu, China \\
}

\begin{document}

\maketitle

\begin{abstract}
  While Large Language Models (LLMs) achieve impressive performance on multi-step reasoning tasks, their reliability is persistently hindered by critical limitations such as unconstrained hallucinations and poor numerical computation. Fundamentally, these issues arise because standard models treat reasoning as a transient, one-off generation process rather than retaining and refining successful procedural logic. To address these challenges, we propose \textbf{eMoT (evolving Memory-of-Thought)}, a unified framework that stabilizes multi-step reasoning by treating reasoning trajectories as dynamic, evolving memories rather than static templates. The framework primarily consists of three interconnected modules: (i) a \textit{memory corrosion} mechanism that reinforces high-utility reasoning structures while gradually decaying less frequent ones; (ii) a \textit{symbolic anchoring} engine that utilizes Python for deterministic computation, much like a human uses a calculator; and (iii) a \textit{consistency-driven refinement} process that aligns neural inference with symbolic outcomes, reducing the accumulation of logical discrepancies. Across multiple reasoning benchmarks, eMoT improves accuracy and solution consistency over standard Chain-of-Thought and structured reasoning baselines.On the traditional task Game of 24, eMoT achieves 100\%  accuracy,surpassing the baseline by up to 17.6\%. Evaluations on mathematical task GSM8K, ASDiv, SVAMP, and MGSM further show consistent gains in multi-step mathematical reasoning. In our evaluation, we achieve superior performance despite utilizing a lightweight backbone model with constrained baseline capabilities. Compared to alternative methods that rely on massively scaled models, our results demonstrate that the performance gains are fundamentally driven by the eMoT framework's reasoning control rather than sheer model size. This suggests that eMoT is not only a powerful booster for smaller-scale models but also offers meaningful methodological references for the development of larger architectures.
\end{abstract}

\section{Introduction}
\begin{figure*}[t]
    \centering
    \includegraphics[width=\textwidth]{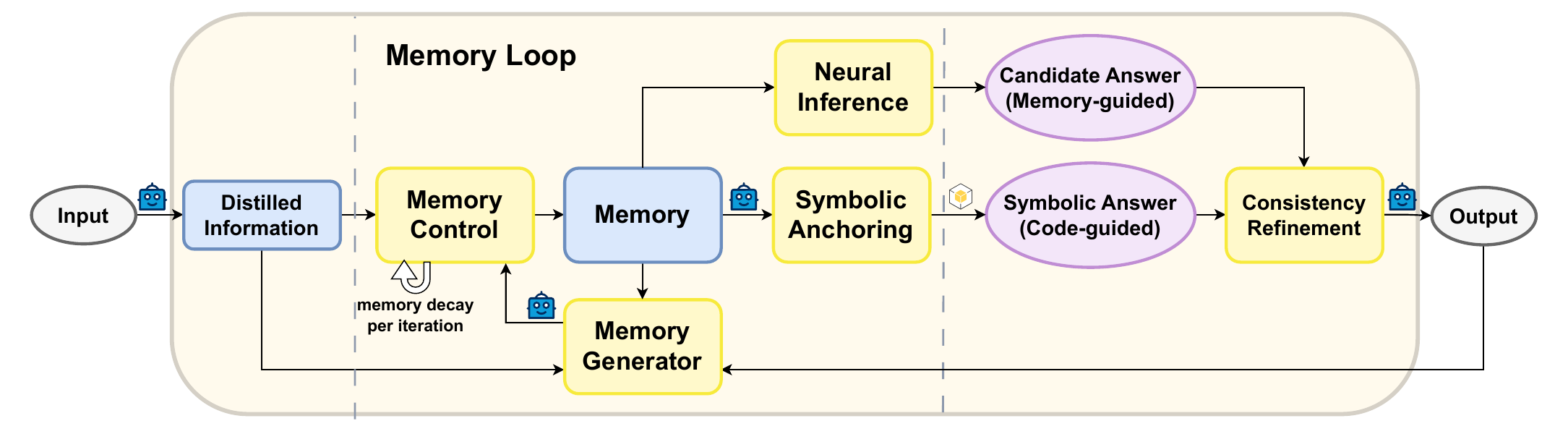}
    \caption{
    \textbf{Overview of eMoT.}
    eMoT mitigates structural drift, intermediate errors, and hallucinated reasoning steps by 
    integrating memory corrosion, symbolic anchoring via Python execution, 
    and consistency-driven refinement.
    Memory control reinforces high-utility reasoning patterns (+$\alpha$) 
    while applying global decay ($(1-\gamma)$) to forget stale or low-utility memories,
    enabling stable and self-evolving multi-step reasoning.
    }
    \label{fig:emot_overview}
\end{figure*}

Large language models (LLMs) have achieved remarkable performance on standard natural language understanding and generation tasks in recent years \citep{brown2020language, chowdhery2023palm, touvron2023llama}. However, these successes largely stem from sophisticated pattern matching and localized context modeling. Multi-step reasoning remains a particularly challenging frontier, as the standard autoregressive generation paradigm struggles to plan ahead, chain intermediate steps accurately, and maintain global logical consistency over extended sequences \citep{rae2021scaling, wang2022self}. While approaches such as Chain-of-Thought (CoT) prompting improve interpretability by producing intermediate reasoning traces \citep{wei2022chain, wang2022self}, these traces are not always reliable explanations of the model’s actual decision process and frequently exhibit unfaithful or inconsistent rationales \citep{lyu2023faithful, turpin2023language}. Such brittleness becomes acutely salient under harder settings where local errors propagate across steps, motivating verification and self-correction mechanisms that explicitly check intermediate claims \citep{dhuliawala2024chain}.

Relatedly, recent work studies how to make reasoning traces more grounded and verifiable by translating them into structured symbolic forms that can be executed or checked \citep{lyu2023faithful}. Deterministic computation or tool-augmented methods, including Python execution engines and programmatic solvers, can guarantee local numerical correctness \citep{gao2023pal, lewkowycz2022solving, chen2022program, schick2023toolformer}, while tool-interactive critique can further improve reliability via iterative checking and revision \citep{gou2023critic}. However, tool use alone does not fully address higher-level structural inconsistencies in long reasoning trajectories. The model may still drift in how it organizes or composes intermediate steps. Other approaches therefore explore richer reasoning structures beyond linear CoT—such as tree-, graph-, or buffer-based inference—to regulate branching, density, and trajectory selection \citep{yao2023tree, besta2024graph, yang2024buffer}. More broadly, research in program synthesis \citep{devlin2017robustfill, austin2021program}, reasoning over knowledge graphs \citep{luo2024reasoning, choudhary2023complex, wu2025graph}, and neuro-symbolic integration \citep{garcez2023neurosymbolic} highlights both the promise and the challenge of combining structured symbolic reasoning with neural models.

Inspired by human cognitive strategies, we propose \textbf{evolving Memory-of-Thought (eMoT)}. As explicitly illustrated in Figure~\ref{fig:emot_overview}, the eMoT framework operates analogously to a temporal state machine, orchestrating a closed-loop architecture driven by an actively updated memory repository. Unlike recent buffer-driven approaches that rely on static templates, eMoT maintains a self-evolving "Memory Loop." Within this loop, retrieved procedural schemas provide specific logical scaffolding rather than pure mathematical abstractions. Each schema is associated with an activation score reflecting its historical utility. Through the Memory Control module depicted in the figure, the system completes a three-stage evolution: retrieving schemas, generating new patterns, and updating scores. High-utility reasoning patterns are continuously reinforced ($+\alpha$), while outdated schemas are subjected to a global exponential decay ($1-\gamma$). This allows the system to operate within a compact, dynamically curated reasoning repository.

Following the memory retrieval phase, the distilled problem information is simultaneously routed into the Symbolic Anchoring module (center of Figure~\ref{fig:emot_overview}). To bridge neural intuition with symbolic precision, eMoT integrates deterministic computation as a logical anchor within the reasoning lifecycle. Unlike Program-Aided Language models (PAL)\citep{gao2023pal}, eMoT treats Python as an external tool for execution rather than an intrinsic part of the reasoning process. Specifically, eMoT generates the complete solution code in a single pass, as opposed to constructing it by stitching together key variables. This approach leverages the proficient coding capabilities of modern LLMs, which significantly minimizes compilation errors. On the other hand, eMoT incorporates an error-handling mechanism that triggers a regeneration upon any execution failure, making the probability of consecutive errors extremely low. By also referencing a "schema" from its memory—which contains validated, functional code templates—eMoT ensures that the generated programs are highly reliable and virtually free of compilation issues.

Furthermore, eMoT does not solely rely on the results derived from Python-based reasoning. Recognizing that complex benchmarks such as GSM-Hard require nuanced logical intuition that symbolic execution alone might fail to capture, the reasoning pathways converge in a dual-stream process, as depicted on the right side of Figure~\ref{fig:emot_overview}. Specifically, the framework deploys a dedicated Neural Inference module to generate a memory-guided Candidate Answer, operating in parallel with the Symbolic Anchoring engine that produces a code-guided Symbolic Answer. If the outputs from both paths are consistent, the result is accepted. In cases of discrepancy, a Consistency Refinement module intervenes to adjudicate between the two and select the optimal final output. This design provides an essential layer of robustness, ensuring that the model delivers accurate results even in the unlikely event of a code-level error.

Experimental results on standard mathematical reasoning datasets, including GSM8K, ASDiv, SVAMP, GSM-Hard, and MGSM, demonstrate that eMoT improves both accuracy and reasoning consistency compared to conventional Chain-of-Thought and structured reasoning baselines.

Our contributions are summarized as follows:

\begin{enumerate}
\item We propose \textbf{evolving Memory-of-Thought (eMoT)}, a novel reasoning framework that models multi-step inference as an evolving repository of reusable procedural schemas, moving beyond static Chain-of-Thought traces and fixed prompt templates.

\item We introduce a synergistic tri-module architecture as the framework's core: a \textit{memory corrosion} mechanism that adaptively reinforces high-utility patterns while decaying obsolete ones; \textit{Python-based symbolic anchoring} for robust deterministic computation; and \textit{consistency-driven refinement} to seamlessly adjudicate divergences between neural intuition and symbolic logic.

\item Comprehensive evaluations demonstrate that eMoT significantly outperforms standard and structured reasoning baselines across diverse mathematical and combinatorial benchmarks. Notably, it achieves near-saturated performance on tasks like Game of 24 and yields substantial accuracy and robustness gains on highly challenging datasets such as GSM-Hard, validating its efficacy in mitigating structural drift and numerical errors.

\end{enumerate}

\section{Related Work}

\paragraph{Evolutions in Multi-Step Reasoning.}
Chain-of-Thought (CoT) prompting enables large language models (LLMs) to decompose complex problems into intermediate reasoning steps, improving interpretability and performance \citep{wei2022chain}. 
Subsequent work explores richer reasoning structures, including tree-based search \citep{yao2023tree}, graph-based reasoning \citep{besta2024graph}, and buffer-based thought management \citep{yang2024buffer}. 
Ensemble strategies such as self-consistency \citep{wang2022self} further improve robustness by aggregating multiple reasoning paths. 
However, recent studies show that long-horizon reasoning remains fragile, with models prone to internal inconsistencies and unreliable intermediate traces \citep{shinn2023reflexion, madaan2023self}.  
In contrast, eMoT maintains a persistent repository of reusable procedural schemas that are reinforced or decayed based on downstream performance.
\paragraph{From Static Scaffolds to Dynamic Schemas.}
Early attempts to stabilize LLM reasoning relied on predefined templates or scratchpads that constrain intermediate reasoning format \citep{nye2021show}. 
More recent work introduces flexible organizational paradigms such as buffer-based reasoning \citep{yang2024buffer} and Chain-of-X frameworks \citep{xia2025beyond}. 
Despite improved flexibility, most prior methods still treat reasoning structures as transient, without maintaining persistent representations of reusable reasoning patterns. 
Unlike prior scaffolding-based methods that constrain reasoning format without maintaining a lifecycle, eMoT explicitly models how reasoning structures are created, reused, reinforced, and eventually discarded.

\paragraph{Symbolic Anchoring and Tool-Augmented Reasoning.}
Integrating deterministic computation has become an effective strategy for reducing arithmetic errors in LLM reasoning \citep{gao2023pal, chen2022program, schick2023toolformer}. 
Recent systems further incorporate execution feedback or tool-based critique to revise model outputs \citep{suris2023vipergpt, gou2023critic}. 
In many existing systems, symbolic execution primarily serves as a post-hoc verifier rather than an explicit constraint on intermediate reasoning steps.
In contrast, eMoT embeds Python-based symbolic execution as a \textit{logical anchor} within the reasoning loop, allowing deterministic outcomes to directly guide intermediate decisions and resolve conflicts between candidate answers.

\paragraph{Procedural Memory and Adaptive Reasoning.}
Memory-augmented neural architectures explore storing and retrieving prior experience to improve generalization \citep{graves2014neural, santoro2016meta}. 
Structured memory systems such as Contextual Memory Trees support scalable and context-aware retrieval \citep{sun2019contextual}. 
While prior work focuses on episodic recall or knowledge retention, eMoT emphasizes \textit{procedural memory} by treating reasoning patterns themselves as reusable and evolving entities. 
Through activation-based reinforcement and controlled decay, eMoT retains high-utility reasoning strategies while gradually phasing out less effective ones.

\paragraph{Broader Hybrid Reasoning Paradigms.}
Hybrid systems combining neural models with symbolic structure have been studied in program synthesis \citep{devlin2017robustfill, austin2021program}, knowledge-graph reasoning \citep{luo2024reasoning, choudhary2023complex, wu2025graph}, and neuro-symbolic learning \citep{garcez2023neurosymbolic}. 
eMoT addresses these limitations by unifying three complementary capabilities within a single framework: evolving procedural memory, symbolic anchoring via Python execution, and consistency-driven refinement for resolving conflicts between candidate answers, forming a cohesive and adaptive hybrid reasoning architecture.

\section{Method}
\subsection{Method Overview}

We present \textbf{eMoT (Evolving Memory-of-Thought)}, a framework that stabilizes multi-step reasoning by combining procedural memory, neural inference, and symbolic execution. Rather than treating reasoning as a transient Chain-of-Thought trace, eMoT maintains an evolving repository of reusable reasoning schemas that guide inference across diverse tasks. Furthermore, Python-based symbolic anchoring is employed as an external verification tool to enhance reasoning accuracy, coupled with robust error-handling strategies to ensure the reliability and executability of the generated code.

\begin{figure*}[!th]
    \centering
    \includegraphics[width=\textwidth]{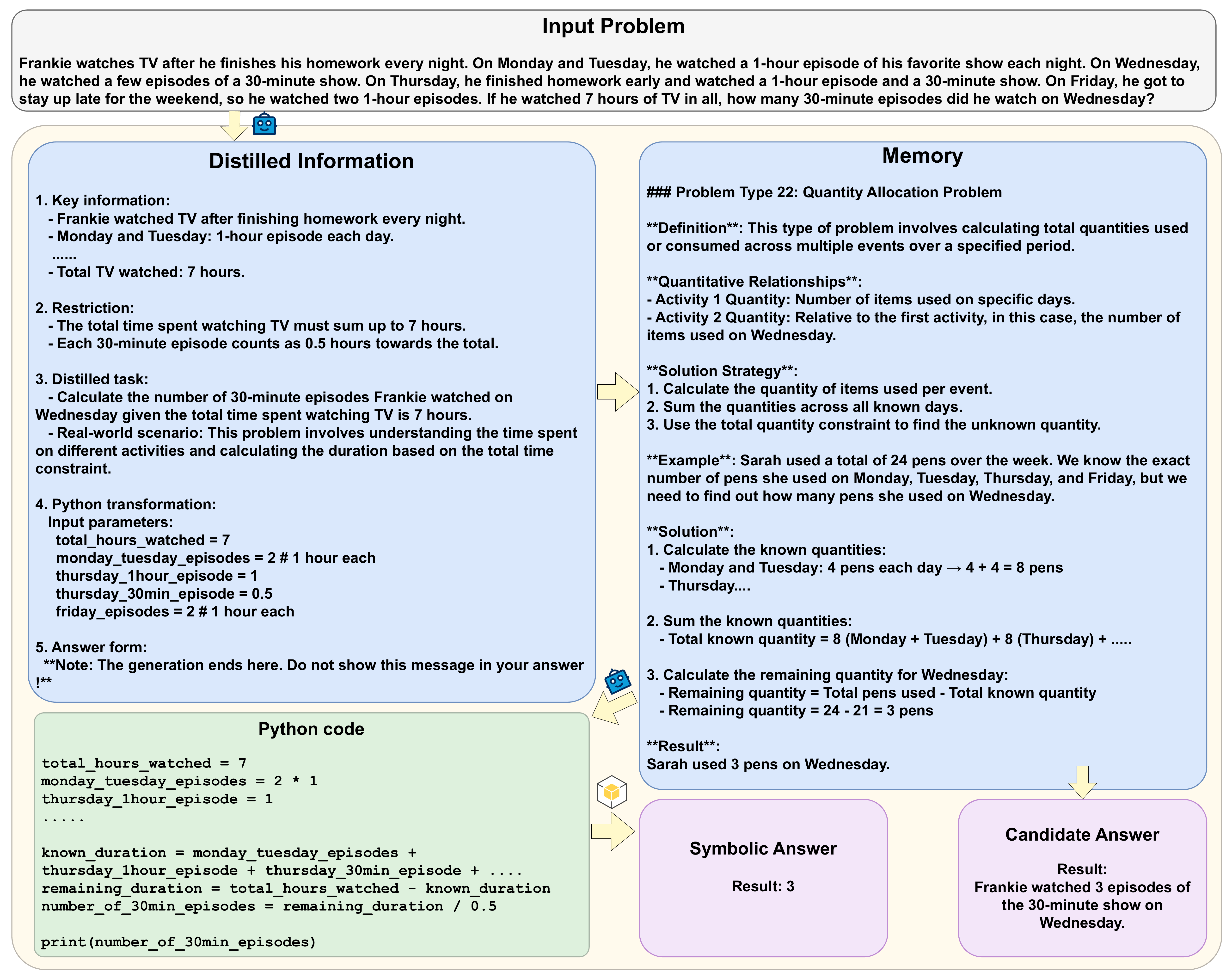}
    \caption{
    \textbf{Worked example of eMoT.}
    The model produces a memory-guided candidate answer and a code-guided symbolic answer,
    which are reconciled through consistency-driven refinement; the outcome updates the evolving memory repository.
    }
    \label{fig:emot_example}
\end{figure*}

Figure~\ref{fig:emot_example} provides a worked example illustrating the end-to-end execution of eMoT on a concrete problem instance. 
Given an input problem $P$, the system first performs \textit{information distillation} to extract a structured representation $I$ that captures key numerical values, constraints, and objectives. 
This representation serves as the shared input to both memory retrieval and symbolic program synthesis.

Based on $I$, eMoT retrieves a set of candidate procedural schemas from the memory repository and selects an activation-weighted schema $s^*$ to scaffold the reasoning process. Conditioned on both $I$ and $s^*$, the language model performs neural inference to generate a structured reasoning trace and a candidate answer. Concurrently, operating as a dual-stream architecture, the symbolic anchoring module synthesizes executable Python code based on $I$ and computes a deterministic result.

If the neural prediction and symbolic output strictly align, the answer is accepted directly. Otherwise, eMoT triggers a consistency-driven refinement step, presenting both candidates to the language model to adjudicate and resolve the discrepancy. Finally, the final optimal answer is utilized to update the memory repository: frequently effective schemas are reinforced $(+\alpha)$, redundant schemas are filtered via similarity-based insertion, and low-utility schemas gradually decay $(1-\gamma)$ and are pruned over time.

Through this closed-loop process, eMoT progressively stabilizes its reasoning structure while retaining the neural flexibility to adapt to novel problem patterns.

\subsection{Mathematical Formulation}

\subsubsection{Information Distillation.} 
Given an input problem $P$ (a natural language problem description), eMoT extracts a structured representation of task-relevant information:
\begin{equation}
I = \text{Distill}(P),
\end{equation}
where $\text{Distill}(\cdot)$ denotes a lightweight information extraction procedure implemented using the language model, and $I$ is a structured representation encoding numerical constants, constraints, and target objectives.
This distilled representation serves as a stable logical substrate for subsequent reasoning.

\subsubsection{Schema Retrieval via RAG.}
\begin{figure*}[h]
    \centering
    \includegraphics[width=0.95\linewidth]{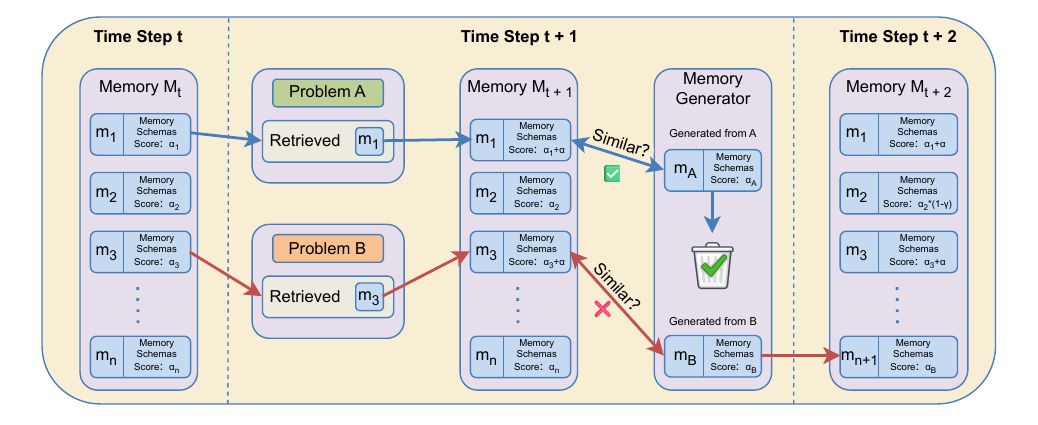}
    \caption{
     At each time step, retrieved schemas are reinforced ($+\alpha$) upon success. Newly generated schemas are evaluated against retrieved ones: redundant candidates are discarded, while novel schemas (e.g., $m_{n+1}$) are appended. Concurrently, unselected schemas undergo global exponential decay ($(1-\gamma)$) to maintain repository compactness.
    }
    \label{fig:memory_lifecycle}
\end{figure*}
Each procedural schema $s_i \in \mathcal{S}$ is stored together with a dense embedding representation computed from its textual description. Given the distilled task information $I$, eMoT retrieves candidate schemas using a retrieval-augmented generation (RAG) pipeline.

Specifically, we encode $I$ into a query embedding using the same sentence-level text encoder $e(\cdot)$ utilized for schema indexing. We compute the cosine similarity between the query embedding and all schema embeddings, selecting the top-$K$ schemas to form the candidate set $\mathcal{S}_K$. The final schema $s^*$ is then selected by maximizing an activation-weighted similarity score:
\begin{equation}
s^* = \arg\max_{s_i \in \mathcal{S}_K} \; a_i \cdot \cos(e(I), e(s_i)),
\end{equation}
where $a_i$ denotes the activation score of schema $s_i$, $K$ is a fixed retrieval hyperparameter, and $\cos(\cdot, \cdot)$ denotes cosine similarity.

This retrieval design decouples semantic similarity estimation from the language model’s generative behavior, ensuring reproducible and controllable schema selection. The language model is only invoked after retrieval to instantiate reasoning conditioned on the selected schema, rather than implicitly judging similarity.

\subsubsection{Logic Scaffolding via Memory Schemas.}

Each retrieved memory schema represents a reusable procedural pattern that specifies how intermediate reasoning steps should be structured. Once a schema $s^*$ is selected, it serves as a \textit{logic scaffold} that constrains the organization of the reasoning trajectory $R = \{r_1, r_2, \dots, r_n\}$ and guides the generation of a neural prediction $y_{\text{LLM}}$.Rather than treating Chain-of-Thought traces as free-form text, eMoT conditions the language model on the selected schema to enforce consistent step ordering, variable usage, and subgoal decomposition. This structured conditioning reduces structural drift by anchoring the reasoning process to previously successful procedural patterns, while still allowing the model to adapt surface realizations to the current input. Consequently, schemas function as persistent structural priors: frequently reused schemas accumulate higher activation scores, thereby increasing their probability of being retrieved in future reasoning episodes.

\subsubsection{Symbolic Anchoring via Deterministic Execution.} 

To explicitly reduce arithmetic and symbolic errors, eMoT delegates executable sub-computations to a deterministic Python-based engine. Specifically, conditioned on the distilled information $I$ and the selected schema $s^*$, the model synthesizes an executable Python program $C$. We formalize the code generation and execution process as:
\begin{equation}
C = \text{GenerateCode}(I, s^*), \quad y_{\text{Symbolic}} = \text{Execute}(C).
\end{equation}

where $\text{GenerateCode}(\cdot)$ denotes a code synthesis function, and $\text{Execute}(\cdot)$ denotes deterministic execution within a Python runtime. The resulting symbolic output $y_{\text{Symbolic}}$ serves as a \textit{logical anchor} providing an external correctness signal. Unlike post-hoc verification, this symbolic execution is intrinsically integrated into the inference loop, directly influencing downstream decision-making by enforcing numerical consistency and strict intermediate variable assignments.

\subsubsection{Consistency-Driven Refinement.}

Given a neural prediction $y_{\text{LLM}}$ and a symbolic result $y_{\text{Symbolic}}$, eMoT performs consistency-driven refinement to resolve any disagreements between neural intuition and deterministic execution. Let $R$ denote the intermediate reasoning trace. The final output $y$ is determined by:
\begin{equation}
y = 
\begin{cases}
y_{\text{LLM}}, & \text{if } y_{\text{LLM}} = y_{\text{Symbolic}}, \\
\text{Refine}(I, s^*, R, y_{\text{Symbolic}}, y_{\text{LLM}}), & \text{otherwise},
\end{cases}
\end{equation}
where $\text{Refine}(\cdot)$ denotes a targeted secondary inference step conditioned on the original problem, the selected schema, the intermediate reasoning trace $R$, and the conflicting candidate answers. Rather than rewriting the entire reasoning chain from scratch, refinement operates strictly at the adjudication level. This mechanism curtails structural drift by preventing locally plausible but globally inconsistent reasoning trajectories from propagating to the final prediction.

\subsubsection{Memory Evolution and Corrosion.} 

\paragraph{Schema Generation and Insertion.}
As illustrated in Figure~\ref{fig:memory_lifecycle}, the memory repository $\mathcal{S}$ evolves across consecutive reasoning episodes. At each time step, eMoT retrieves a memory schema $s^*$ to structure the current reasoning process. Upon producing a valid solution, the system synthesizes a new candidate procedural schema $s_{\text{new}}$ abstracted from the current reasoning trace.

To prevent redundant memory bloat, $s_{\text{new}}$ is evaluated against the originally selected schema $s^*$. If the two schemas exhibit high semantic similarity, the candidate is discarded; conversely, if it represents a novel strategy, it is inserted into the repository as a new reusable pattern. This dynamic insertion expands reasoning coverage without duplicating previously stored structures.

\paragraph{Memory Corrosion and Purge.}
Each stored schema $s_j \in \mathcal{S}$ is intrinsically tied to an activation score $a_j$, which reflects its historical utility. Let $\alpha$ denote the reinforcement increment, $\gamma$ the corrosion (decay) rate, and $\theta$ the baseline purge threshold. Schemas utilized during a successful episode receive positive reinforcement, while the entire repository simultaneously undergoes global exponential decay:
\begin{equation}
a_j \leftarrow 
\begin{cases} 
a_j + \alpha, & \text{if selected}, \\
(1-\gamma)a_j, & \text{otherwise}.
\end{cases}
\end{equation}

Subsequently, schemas whose activation scores naturally corrode below the purge threshold $\theta$ undergo automated garbage collection and are phased out of the active repository:
\begin{equation}
\mathcal{S} \leftarrow \left\{ (s_j, a_j) \in \mathcal{S} \;\middle|\; a_j \ge \theta \right\}.
\label{eq:memory_purge}
\end{equation}

This dual corrosion-and-purge mechanism ensures the memory space remains remarkably compact. By selectively reinforcing high-utility procedural patterns and systematically decaying obsolete schemas until they are naturally phased out, eMoT facilitates highly efficient, continual adaptation while maintaining strict bounds on memory footprint.

\section{Experiments}

\subsection{Experimental Setup}

\paragraph{Datasets.} 
We evaluate eMoT across a comprehensive suite of arithmetic, algebraic, and combinatorial reasoning benchmarks. The mathematical evaluation incorporates GSM8K \citep{cobbe2021training}, ASDiv \citep{miao2020diverse}, SVAMP \citep{patel2021nlp}, and MGSM \citep{cobbe2021training}, which collectively assess multi-step deductive capabilities under diverse linguistic and structural complexities. Furthermore, we report performance on combinatorial Chain-of-Thought benchmarks (Game of 24, WordSorting, and Checkmate) adopting the evaluation protocols established by Tree-of-Thought \citep{yao2023tree}. To specifically stress-test long-horizon reasoning, we also include GSM-Hard, a highly challenging variant of the GSM8K dataset.

\paragraph{Models and Baselines.} 
Throughout our experiments, we employ Qwen-32B as the primary foundation model. We rigorously benchmark eMoT against a representative set of reasoning paradigms: standard Chain-of-Thought (CoT) prompting \citep{wei2022chain}, Tree-of-Thought (ToT) \citep{yao2023tree}, Buffer-of-Thoughts (BoT) \citep{yang2024buffer}, and Program-Aided Language Models (PaL) \citep{gao2023pal}. Performance metrics for GPT-4 and Codex-based methodologies are cited directly from their original publications under matched evaluation settings. Additionally, we construct ablation variants of eMoT to selectively isolate and evaluate the impact of memory control, symbolic execution, and the consistency-driven refinement module.

\paragraph{Metrics.} 
We report the exact-match accuracy as the primary evaluation metric across all arithmetic and algebraic datasets. For the combinatorial and logic-oriented tasks (Game of 24, WordSorting, Checkmate), correctness is strictly determined using their standard task-specific success criteria.

\subsection{Main Results}

\begin{table*}[ht]
\centering
\resizebox{0.98\textwidth}{!}{
\begin{tabular}{l l c c c c c c c c}
\toprule
\textbf{Method} & \textbf{Models} &
\textbf{ASDiv} & \textbf{GSM8K} & \textbf{GSM-Hard} &
\textbf{MGSM} & \textbf{SVAMP} &
\textbf{Game24} & \textbf{WordSort} & \textbf{Checkmate} \\
\midrule

Qwen-32B (Direct) & Qwen-32B
& 0.833 & 0.395 & 0.159 & 0.480 & 0.830
& -- & -- & -- \\

BoT & GPT-4
& 0.928 & 0.933 & 0.620 & 0.893 & 0.913
& 0.824 & \textbf{1.000} & 0.864 \\

ToT & GPT-4
& 0.573 & 0.341 & 0.244 & 0.328 & 0.600
& 0.740 & 0.964 & 0.492 \\

PaL & Codex (175B) 
& 0.796 & 0.720 & 0.612 & 0.736 & 0.794
& -- & -- & -- \\
{eMoT (Ours)} & Qwen-32B
& \textbf{0.944} & \textbf{0.934} & \textbf{0.715} & \textbf{0.944} & \textbf{0.940}
& \textbf{1.000} & 0.968 & \textbf{0.900} \\
\bottomrule
\end{tabular}
}
\caption{
\textbf{Main results on multi-step reasoning benchmarks.}
All methods are evaluated using the same backbone (Qwen-32B) unless otherwise specified.
Results for GPT-4 and CODEX are taken from prior work under their reported evaluation protocols.
“-” denotes that the tasks in this modality are not supported by the model.
}
\label{tab:main_results}
\end{table*}
As summarized in Table~\ref{tab:main_results}, eMoT consistently outperforms direct prompting with Qwen-32B across all arithmetic and multilingual benchmarks. On GSM8K, accuracy surges from 0.395 (Direct) to 0.934, and crucially, on GSM-Hard from 0.159 to 0.715. These substantial gains highlight the framework's efficacy in handling problems that require deeper, sustained multi-step reasoning. Similar improvements are observed on ASDiv (0.833 $\rightarrow$ 0.944), MGSM (0.480 $\rightarrow$ 0.944), and SVAMP (0.830 $\rightarrow$ 0.940), demonstrating that eMoT generalizes robustly across diverse mathematical reasoning formats.

Compared with the structured reasoning baselines reported in Table~\ref{tab:main_results}, eMoT exhibits remarkable competitiveness despite operating on a significantly smaller backbone model. While BoT and ToT rely on the massive GPT-4, eMoT achieves superior or comparable accuracy on most datasets using Qwen-32B. This advantage is particularly pronounced on GSM-Hard (0.715 for eMoT vs. 0.620 for BoT and 0.244 for ToT), indicating that the observed performance leaps are fundamentally driven by improved reasoning control rather than mere parameter scaling.

Furthermore, we observe systematic performance disparities across different reasoning paradigms. BoT performs strongly on arithmetic-heavy datasets such as GSM8K and ASDiv but struggles on combinatorial and logic-oriented tasks (e.g., Checkmate). ToT enhances exploration through tree-based search but remains highly susceptible to cascading errors as reasoning depth increases, which is reflected in its weaker performance on GSM-Hard and MGSM. In contrast, PaL benefits from deterministic program execution on arithmetic datasets but lacks native extensibility to combinatorial tasks like Game of 24.

Beyond aggregate mathematical accuracy, Table~\ref{tab:main_results} reveals that eMoT achieves near-saturated performance on combinatorial benchmarks, reaching a perfect solve rate of 1.000 on Game of 24, alongside 0.968 on WordSorting and 0.900 on Checkmate. These results validate that the integration of procedural memory with symbolic anchoring effectively supports both expansive structured search and precise intermediate computation. Across all evaluated datasets, eMoT yields the most pronounced relative gains on tasks requiring extended reasoning chains, which aligns perfectly with its core capacity to reuse and stabilize effective decomposition patterns under fixed decoding constraints.

\subsection{Analysis}
\paragraph{Mitigating Structural Drift.}
We characterize structural drift as a critical failure mode in which intermediate reasoning gradually departs from global task constraints, even when individual steps remain locally plausible. This phenomenon is particularly pronounced in long-horizon reasoning, where minor early errors compound into globally inconsistent conclusions.

In eMoT, structural drift is systematically mitigated by explicitly coupling memory-guided inference with deterministic Python execution. Rather than relying solely on latent chain-of-thought signals, the framework cross-validates neural predictions against an external symbolic anchor. When a discrepancy arises, eMoT triggers a secondary adjudication step that forces the model to reconcile conflicting candidates, effectively converting silent reasoning errors into explicit decision conflicts.

Importantly, this refinement operates at the decision level rather than attempting to rewrite intermediate traces, which are often noisy and unreliable. This architectural design constrains divergence without over-regularizing the underlying reasoning process, enabling eMoT to suppress error propagation while preserving the flexibility required for complex multi-step inference.

\paragraph{Schema Repository Dynamics.}
The schema repository evolves through a reinforcement-and-decay mechanism. Frequently selected schemas accumulate activation scores, gradually converging into a compact foundation of canonical decomposition patterns. Conversely, rarely retrieved schemas undergo multiplicative decay. This dynamic induces an implicit curriculum that systematically marginalizes idiosyncratic templates, functioning as a continuous, automated garbage collection that stabilizes long-term reasoning without disruptive manual pruning.Crucially, corrosion parameters (e.g., decay rate $\gamma$) provide the essential evolutionary selection pressure to prevent retrieval poisoning from obsolete schemas. While fundamental to maintaining high reasoning accuracy, eMoT is not brittle to exact hyperparameter tuning. The framework exhibits a dynamic convergence property: operating within a reasonable basin (e.g., $\gamma \in [0.001, 0.01]$), the repository robustly converges toward the same high-utility structures. Consequently, exhaustive grid search is unnecessary; the core contribution lies in this self-optimizing evolutionary architecture, which organically drives the system toward optimal reasoning patterns without requiring fragile calibration.

\subsection{Ablation Study}

To systematically isolate the contribution of each component within the eMoT framework, we conduct rigorous ablation experiments by selectively disabling memory retrieval, symbolic execution, or consistency-driven refinement, while strictly maintaining the backbone model (Qwen-32B) and decoding configurations.
\begin{table}[ht]
    \centering
    \begin{tabular}{lcccccc}
        \toprule
        \textbf{Model} &  \textbf{ASDiv} & \textbf{GSM8K} & \textbf{GSM-Hard} & \textbf{MGSM} & \textbf{SVAMP} \\
        \midrule
        eMoT & \textbf{0.944} & \textbf{0.934} & \textbf{0.715} & \textbf{0.944} & \textbf{0.940} \\
        \midrule
        w/o Mem  & 0.921 & 0.928 & 0.660 & 0.916 & 0.923 \\
        w/o Sym   & 0.903 & 0.916 & 0.656 & 0.908 & 0.898 \\
        w/o Refine  & 0.935 & 0.929 & 0.694 & 0.938 & 0.929 \\
        \bottomrule
    \end{tabular}
    \caption{Ablation results of eMoT on arithmetic reasoning benchmarks.}
    \label{tab:ablation}
\end{table}

Removing the procedural memory module (w/o Mem) leads to consistent performance degradation across all datasets, with the most pronounced absolute drop observed on GSM-Hard (0.715 $\rightarrow$ 0.660). This pattern suggests that procedural memory contributes far more than marginal accuracy improvements: it provides essential structural priors that stabilize reasoning under escalating depth and difficulty, scenarios where naive step-by-step prompting becomes highly susceptible to compounding errors.

Disabling Python-based symbolic execution (w/o Sym) incurs the most severe performance penalties on arithmetic-intensive benchmarks such as ASDiv, GSM8K, and SVAMP. These results confirm that deterministic computation functions primarily as a critical reliability anchor, fortifying the model's robustness against numerical errors and preventing cascading arithmetic mistakes from corrupting the downstream reasoning trajectory.

Omitting the consistency-driven refinement module (w/o Refine) yields a more subtle, yet strictly systematic, performance decline, particularly on complex instances. This indicates that refinement does not merely patch isolated arithmetic mistakes; rather, it operates as a sophisticated decision-level safeguard that resolves conceptual conflicts between neural heuristics and symbolic outcomes. By transforming latent reasoning inconsistencies into explicit adjudication steps, the refinement module successfully recovers a critical subset of predictions that would otherwise fail due to early-stage logical divergence.

\section{Conclusion}

We introduced eMoT (evolving Memory-of-Thought), a reasoning framework that advances multi-step inference by replacing transient chain-of-thought traces with an evolving repository of reusable procedural schemas. By integrating persistent memory, deterministic symbolic anchoring, and consistency-driven refinement, eMoT significantly enhances long-horizon reasoning stability.

Evaluations across diverse mathematical and combinatorial benchmarks demonstrate that eMoT consistently outperforms existing structured reasoning baselines, achieving substantial gains on challenging instances like GSM-Hard. Our analysis confirms the synergistic impact of its core components: memory promotes effective pattern reuse, symbolic execution curtails numerical error propagation, and refinement reliably adjudicates neural-symbolic divergences.

While demonstrating strong empirical performance, eMoT's multi-module routing inherently incurs higher token consumption than standard zero-shot CoT. Nevertheless, it provides a highly favorable efficiency-reliability trade-off compared to brute-force test-time scaling (e.g., extensive Self-Consistency voting). Future research will focus on extending generalizability to competition-level domains (e.g., AIME, GPQA), replacing single-pass code synthesis with iterative auto-debugging loops (akin to ReTool), and developing learnable memory management strategies for broader open-ended reasoning tasks.

\medskip

{
\small
\bibliographystyle{unsrt}
\bibliography{new_cite}
}


\clearpage
\appendix
\section*{Supplementary Material}

\subsection*{A. Qualitative Examples}

\paragraph{A.1 A Correct Example from ASDiv: Comparative Quantity.}
We illustrate a representative instance where eMoT solves a comparative quantity problem via structured distillation, schema retrieval, and symbolic execution.

\begin{figure}[th]
  \centering
  \includegraphics[width=0.95\linewidth]{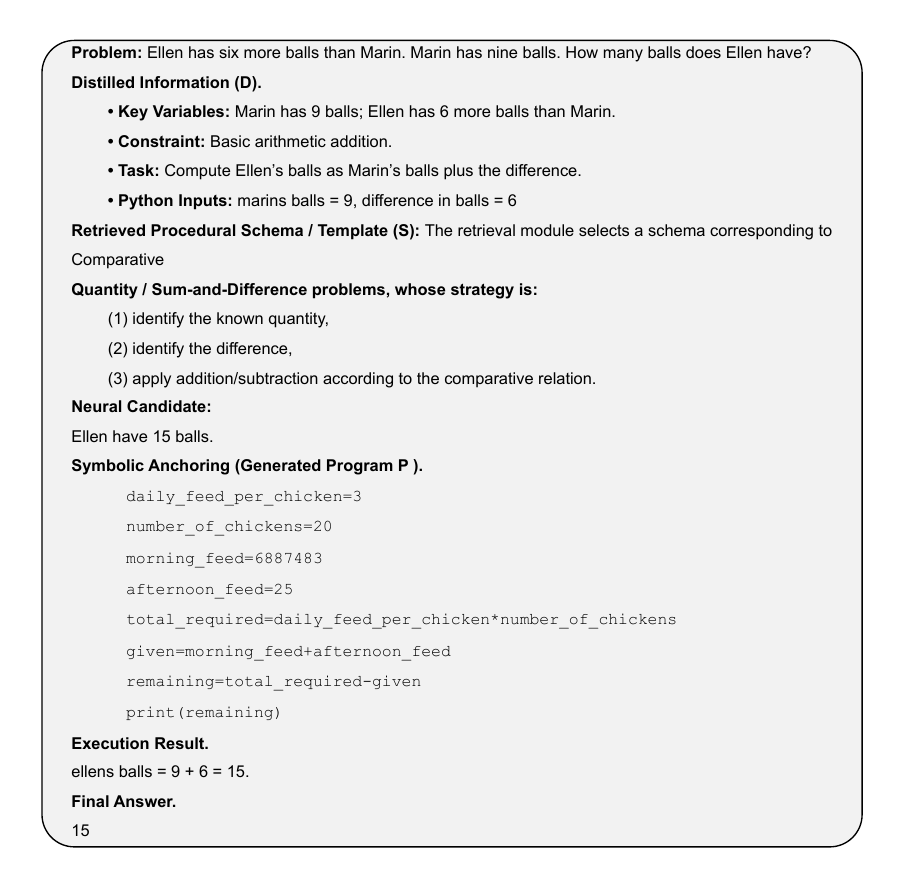}
  \caption{
  \textbf{End-to-end reasoning trace for a comparative quantity problem (ASDiv).}
  }
  \label{fig:asdiv-example1}
\end{figure}

\noindent\textbf{Mechanistic Insight.}
This example highlights how eMoT elegantly separates problem parsing (distillation) from deterministic computation (symbolic anchoring). Meanwhile, the procedural memory provides an abstract solution pattern (comparative quantity) without overfitting to or copying instance-specific values.

\paragraph{A.2 A Boundary-Condition Example from GSM-Hard: Extreme Numerical Values Induce Neural Plausibility Drift.}
We present a representative example where extreme numeric magnitudes cause the neural reasoning branch to introduce unintended commonsense-based corrections, while the symbolic branch preserves mathematically correct computation.

\begin{figure}[h!]
  \centering
  \includegraphics[width=0.95\linewidth]{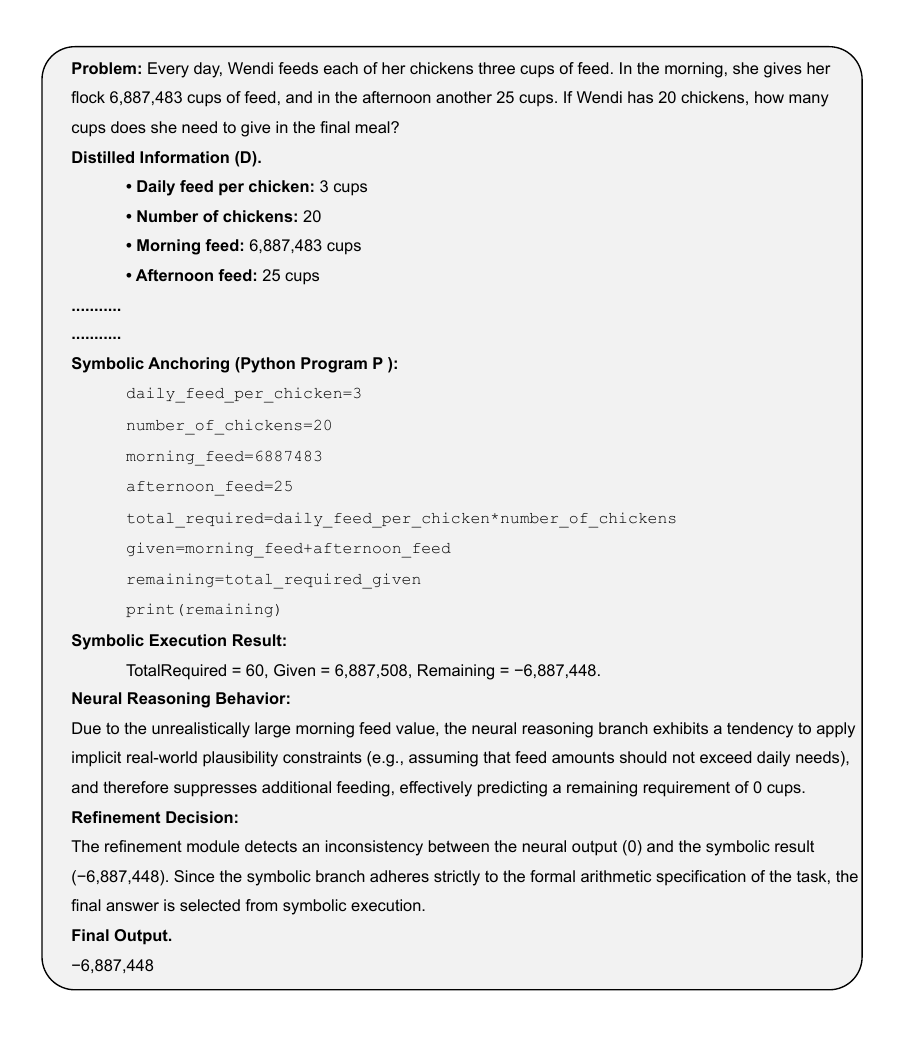}
  \caption{
  \textbf{Boundary-condition example from GSM-Hard highlighting formal correctness vs human plausibility.}
  }
  \label{fig:gsmhard-example}
\end{figure}

\noindent\textbf{Mechanistic Insight.}
This case demonstrates that extreme numeric regimes can trigger \textit{commonsense plausibility drift} in neural reasoning, whereas symbolic anchoring provides a distribution-robust computation pathway. The refinement step serves as a critical safeguard, overriding heuristic neural corrections in favor of formally correct symbolic outcomes. We note that benchmarks such as GSM-Hard intentionally include edge-case arithmetic scenarios that stress-test formal numerical reasoning under unrealistic conditions. In such settings, mathematically correct answers may naturally conflict with human commonsense expectations. This example should therefore be interpreted as a boundary-condition evaluation of symbolic correctness, underscoring eMoT’s capacity to prioritize formal task specifications over misleading human cognitive priors.

\paragraph{A.3 A Failure-Recovery Example from MGSM: Symbolic Code Execution Error and Fallback Recovery.}
We illustrate a representative failure mode where the symbolic anchoring branch generates syntactically invalid Python code (e.g., a variable-name mismatch). In such cases, eMoT gracefully falls back to the neural reasoning answer, preserving a valid final output.

\begin{figure}[h!]
  \centering
  \includegraphics[width=0.95\linewidth]{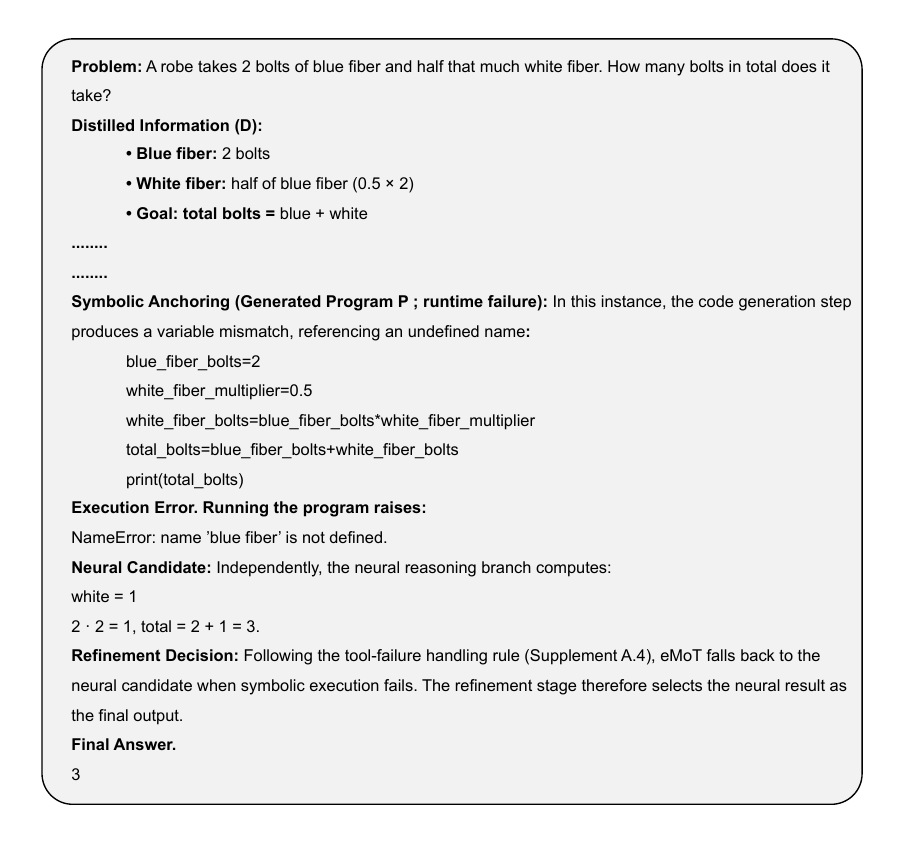}
  \caption{
  \textbf{Symbolic code execution failure with neural fallback and refinement recovery.}
  }
  \label{fig:mgsm-example}
\end{figure}

\noindent\textbf{Mechanistic Insight.}
This case highlights a practical limitation of tool-augmented reasoning: even when the underlying mathematical structure is sound, symbolic anchoring may fail due to low-level implementation issues, such as variable binding errors. eMoT mitigates such vulnerabilities via a robust fallback policy, ensuring system resilience without discarding the problem instance. Crucially, unlike traditional multi-step Chain-of-Thought reasoning—where a single early hallucination irreversibly corrupts the entire downstream solution—symbolic anchoring in eMoT typically involves a single-pass code synthesis. This flat generation structure drastically reduces the probability of cascading logical failures. When synthesis errors do occur, they are isolated to the code generation step itself. The refinement module acts as a resilient decision-level safeguard, detecting these execution failures and seamlessly routing the system to an alternative valid neural solution.

\clearpage
\subsection*{B. System Prompts for eMoT}

\paragraph{B.1. Prompt for Information Distillation}
\begin{mdframed}[backgroundcolor=gray!10, roundcorner=5pt]
\small
Please distill the information following the format below and cease response after the output of the distilled information.

\textbf{Meta distiller Respond:}\\
\textbf{Distilled Information:}\\
1. Key information:\\
2. Restriction: (It should be noted that the answer should strictly follow real-world rules such as in arithmetic equations, the Priority of operators, the need for parentheses, etc. So according to the distilled information, emphasize the real-world rules that need to be followed within the problem.)\\
3. Distilled task:\\
4. Python transformation:\\
(Optional, skip when Python tag is Not for Python) Input parameters: (The names of each variable should be clear and not confusing, and correspond to the entity names in the problem)\\
variable1\_name = x\\
variable2\_name = y\\
variableN\_name = z\\
5. Answer form: (Optional, skip when there is no specific answer form)
\end{mdframed}

\paragraph{B.2. Prompt for Symbolic Anchoring}
\begin{mdframed}[backgroundcolor=gray!10, roundcorner=5pt]
\small
You are an expert in mathematical problem solving and symbolic reasoning.
You will receive:\\
1. \textbf{Distilled Information:} a structured summary of a math word problem (variables, quantities, and task goals).\\
2. \textbf{Memory (RAG):} a collection of generalized reasoning Memory for different math problem types.

Your task:\\
1. Use this memory to generate a \textbf{fully executable Python code snippet} that computes the correct numeric answer.\\
2. The generated code must:\\
- Use \textbf{only} the variables and numeric values explicitly provided in the distilled information.\\
- Be \textbf{fully self-contained} (no \texttt{input()}, \texttt{import}, \texttt{random}, or external calls).\\
- Be \textbf{clean, minimal, and deterministic} - only arithmetic, logic, and assignment operations are allowed.\\
- End with a single \texttt{print(<final\_answer\_variable\_or\_expression>)} statement.\\
3. Do \textbf{not} include explanations, reasoning, or comments.\\
4. The \textbf{entire output} must be wrapped strictly in triple backticks with a \texttt{python} tag. Do not output anything except the Python code block.
\end{mdframed}

\paragraph{B.3. Prompt for Consistency-Driven Refinement}
\begin{mdframed}[backgroundcolor=gray!10, roundcorner=5pt]
\small
You are an expert reasoning analyst. The following is a distilled thought template and original reasoning result.

\textbf{Task:}\\
1. Analyze the quantitative relationships and solution strategy from the thought template.\\
2. Perform correct calculations based on the relationships.\\
3. Compare with the original reasoning result and template result.\\
4. Output only the correct numeric answer.

\textbf{Final Refined Result:} <your numeric answer>
\end{mdframed}


\newpage
\input{checklist.tex}

\end{document}

%% file: checklist.tex
\section*{NeurIPS Paper Checklist}

The checklist is designed to encourage best practices for responsible machine learning research, addressing issues of reproducibility, transparency, research ethics, and societal impact. Do not remove the checklist: {\bf The papers not including the checklist will be desk rejected.} The checklist should follow the references and follow the (optional) supplemental material.  The checklist does NOT count towards the page
limit. 

Please read the checklist guidelines carefully for information on how to answer these questions. For each question in the checklist:
\begin{itemize}
    \item You should answer \answerYes{}, \answerNo{}, or \answerNA{}.
    \item \answerNA{} means either that the question is Not Applicable for that particular paper or the relevant information is Not Available.
    \item Please provide a short (1--2 sentence) justification right after your answer (even for \answerNA). 
\end{itemize}

{\bf The checklist answers are an integral part of your paper submission.} They are visible to the reviewers, area chairs, senior area chairs, and ethics reviewers. You will also be asked to include it (after eventual revisions) with the final version of your paper, and its final version will be published with the paper.

The reviewers of your paper will be asked to use the checklist as one of the factors in their evaluation. While \answerYes{} is generally preferable to \answerNo{}, it is perfectly acceptable to answer \answerNo{} provided a proper justification is given (e.g., error bars are not reported because it would be too computationally expensive'' or ``we were unable to find the license for the dataset we used''). In general, answering \answerNo{} or \answerNA{} is not grounds for rejection. While the questions are phrased in a binary way, we acknowledge that the true answer is often more nuanced, so please just use your best judgment and write a justification to elaborate. All supporting evidence can appear either in the main paper or the supplemental material, provided in appendix. If you answer \answerYes{} to a question, in the justification please point to the section(s) where related material for the question can be found.

IMPORTANT, please:
\begin{itemize}
    \item {\bf Delete this instruction block, but keep the section heading ``NeurIPS Paper Checklist"},
    \item  {\bf Keep the checklist subsection headings, questions/answers and guidelines below.}
    \item {\bf Do not modify the questions and only use the provided macros for your answers}.
\end{itemize}


\begin{enumerate}

\item {\bf Claims}
    \item[] Question: Do the main claims made in the abstract and introduction accurately reflect the paper's contributions and scope?
    \item[] Answer: \answerYes{} 
    \item[] Justification: The abstract and introduction clearly state the framework's components (memory corrosion, symbolic anchoring, refinement) and explicitly outline the claims regarding performance improvements on mathematical reasoning tasks, which are directly supported by the empirical evaluations in Section 4.

\item {\bf Limitations}
    \item[] Question: Does the paper discuss the limitations of the work performed by the authors?
    \item[] Answer: \answerYes{} 
    \item[] Justification: We explicitly discuss the limitations of our framework—including higher token consumption, multi-module latency, and sensitivity to initial information extraction—in the final paragraph of the Conclusion section, alongside corresponding avenues for future research.

\item {\bf Theory assumptions and proofs}
    \item[] Question: For each theoretical result, does the paper provide the full set of assumptions and a complete (and correct) proof?
    \item[] Answer: \answerNA{} 
    \item[] Justification: This paper focuses on an empirical systems architecture and inference methodology for LLMs. It evaluates empirical performance and does not present formal mathematical theorems, theoretical bounds, or proofs.

    \item {\bf Experimental result reproducibility}
    \item[] Question: Does the paper fully disclose all the information needed to reproduce the main experimental results of the paper to the extent that it affects the main claims and/or conclusions of the paper (regardless of whether the code and data are provided or not)?
    \item[] Answer: \answerYes{} 
    \item[] Justification: We provide comprehensive details of the experimental setup in Section 4.1, including datasets, baselines, and the backbone model. Furthermore, the complete system prompts and step-by-step examples necessary to replicate the inference pipeline are explicitly provided in Appendix A and B.

\item {\bf Open access to data and code}
    \item[] Question: Does the paper provide open access to the data and code, with sufficient instructions to faithfully reproduce the main experimental results, as described in supplemental material?
    \item[] Answer: \answerNo{} 
    \item[] Justification: We have provided the complete system prompts, schematic logic, and qualitative examples in the Appendix to ensure algorithmic transparency and reproducibility. We plan to open-source the full codebase, including data processing and execution environments, upon the acceptance of this paper.

\item {\bf Experimental setting/details}
    \item[] Question: Does the paper specify all the training and test details (e.g., data splits, hyperparameters, how they were chosen, type of optimizer) necessary to understand the results?
    \item[] Answer: \answerYes{} 
    \item[] Justification: Datasets, evaluation metrics, baseline approaches, and the backbone model (Qwen-32B) utilized for our inference framework are fully described in Section 4.1. The hyperparameter robustness (e.g., decay rate) is discussed in Section 4.3.

\item {\bf Experiment statistical significance}
    \item[] Question: Does the paper report error bars suitably and correctly defined or other appropriate information about the statistical significance of the experiments?
    \item[] Answer: \answerNo{} 
    \item[] Justification: Following standard practice for evaluation on fixed mathematical reasoning benchmarks (e.g., GSM8K, Game of 24), we report exact-match accuracy over the entire test sets rather than running multiple random seeds to compute error bars, given the high inference costs associated with multi-step LLM generations.

\item {\bf Experiments compute resources}
    \item[] Question: For each experiment, does the paper provide sufficient information on the computer resources (type of compute workers, memory, time of execution) needed to reproduce the experiments?
    \item[] Answer: \answerNo{} 
    \item[] Justification: While we specify the foundation model (Qwen-32B) and extensively discuss inference efficiency trade-offs, we currently omit explicit hardware details (e.g., exact GPU types, total execution hours). We will include comprehensive compute resource specifications in the camera-ready appendix.
    
\item {\bf Code of ethics}
    \item[] Question: Does the research conducted in the paper conform, in every respect, with the NeurIPS Code of Ethics \url{https://neurips.cc/public/EthicsGuidelines}?
    \item[] Answer: \answerYes{} 
    \item[] Justification: The research conducted in this paper fully conforms, in every respect, with the NeurIPS Code of Ethics.

\item {\bf Broader impacts}
    \item[] Question: Does the paper discuss both potential positive societal impacts and negative societal impacts of the work performed?
    \item[] Answer: \answerNo{} 
    \item[] Justification: Our work focuses on a general inference framework (eMoT) for improving mathematical and logic reasoning in LLMs. We do not foresee direct negative societal impacts stemming from this specific methodology, beyond the generic dual-use nature of foundational language models. Therefore, a dedicated broader impacts section is omitted.
    
\item {\bf Safeguards}
    \item[] Question: Does the paper describe safeguards that have been put in place for responsible release of data or models that have a high risk for misuse (e.g., pre-trained language models, image generators, or scraped datasets)?
    \item[] Answer: \answerNA{} 
    \item[] Justification: We do not release new high-risk datasets or foundational pre-trained models. Our contribution is an inference-time framework applied to existing, publicly available open-source models (e.g., Qwen-32B).

\item {\bf Licenses for existing assets}
    \item[] Question: Are the creators or original owners of assets (e.g., code, data, models), used in the paper, properly credited and are the license and terms of use explicitly mentioned and properly respected?
    \item[] Answer: \answerYes{} 
    \item[] Justification: All datasets used in our evaluation (e.g., GSM8K, ASDiv, SVAMP, MGSM) and the backbone model (Qwen-32B) are open-source assets. They are properly credited and cited in the experimental setup (Section 4.1).

\item {\bf New assets}
    \item[] Question: Are new assets introduced in the paper well documented and is the documentation provided alongside the assets?
    \item[] Answer: \answerYes{} 
    \item[] Justification: The new asset introduced in this work (the eMoT framework codebase and its corresponding evolving schema repository) will be released under a standard open-source license upon publication, accompanied by comprehensive documentation and usage instructions in the public repository.

\item {\bf Crowdsourcing and research with human subjects}
    \item[] Question: For crowdsourcing experiments and research with human subjects, does the paper include the full text of instructions given to participants and screenshots, if applicable, as well as details about compensation (if any)? 
    \item[] Answer: \answerNA{} 
    \item[] Justification: The research does not involve human subjects or crowdsourcing experiments.

\item {\bf Institutional review board (IRB) approvals or equivalent for research with human subjects}
    \item[] Question: Does the paper describe potential risks incurred by study participants, whether such risks were disclosed to the subjects, and whether Institutional Review Board (IRB) approvals (or an equivalent approval/review based on the requirements of your country or institution) were obtained?
    \item[] Answer: \answerNA{} 
    \item[] Justification: The research does not involve human subjects or crowdsourcing experiments.

\item {\bf Declaration of LLM usage}
    \item[] Question: Does the paper describe the usage of LLMs if it is an important, original, or non-standard component of the core methods in this research? Note that if the LLM is used only for writing, editing, or formatting purposes and does \emph{not} impact the core methodology, scientific rigor, or originality of the research, declaration is not required.
    \item[] Answer: \answerYes{} 
    \item[] Justification: The core methodology of this research inherently revolves around orchestrating Large Language Models. The use of Qwen-32B as the backbone for inference, memory generation, code synthesis, and refinement is extensively detailed throughout the main text and the experimental setup (Section 4.1).

\end{enumerate}